\documentclass[10pt,twocolumn,letterpaper]{article}

\usepackage[pagenumbers]{cvpr} 

\definecolor{cvprblue}{rgb}{0.21,0.49,0.74}
\usepackage[pagebackref,breaklinks,colorlinks,citecolor=cvprblue]{hyperref}
\usepackage{algpseudocode}
\usepackage{algorithm}
\usepackage{comment}
\newcommand{\mname}[1]{\textcolor{black}{PreciseCam}} 
\newcommand{\supplem}[1]{\textcolor{black}{#1}}

\def\paperID{13640} 
\def\confName{CVPR}
\def\confYear{2025}

\title{\mname{}: Precise Camera Control for Text-to-Image Generation}

\author{Edurne Bernal-Berdun\\
Universidad de Zaragoza, I3A\\
{\tt\small edurnebernal@unizar.es}
\and
Ana Serrano\\
Universidad de Zaragoza, I3A\\
{\tt\small anase@unizar.es}
\and
Belen Masia\\
Universidad de Zaragoza, I3A\\
{\tt\small bmasia@unizar.es}
\and
Matheus Gadelha\\
Adobe Research\\
{\tt\small gadelha@adobe.com}
\and
Yannick Hold-Geoffroy\\
Adobe Research\\
{\tt\small holdgeof@adobe.com}
\and
Xin Sun\\
Adobe Research\\
{\tt\small xinsun@adobe.com}
\and
Diego Gutierrez\\
Universidad de Zaragoza, I3A\\
{\tt\small diegog@unizar.es}
}

\begin{document}
\twocolumn[{%
\renewcommand\twocolumn[1][]{#1}%
\maketitle
\begin{center}
    \centering
    \captionsetup{type=figure}
    \includegraphics[width=\textwidth]{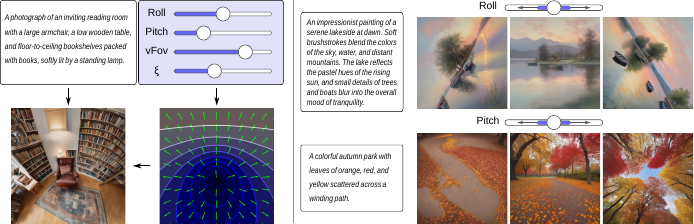}
    \captionof{figure}{Our approach enhances the artistic expression of text-to-image generative models by incorporating precise control over camera angles and lens distortion effects. \textbf{Left:} Our input consists of a standard text prompt along with extrinsic (roll and pitch) and intrinsic (vertical field of view and distortion $\xi$) camera parameters, which are translated into a suitable and efficient representation for learning camera views. \textbf{Right}: Examples varying roll (top) and pitch (bottom) with the same prompt, while keeping the remaining camera parameters fixed.}
\end{center}%

}]

\begin{abstract}
Images as an artistic medium often rely on specific camera angles and lens distortions to convey ideas or emotions; however, such precise control is missing in current text-to-image models. We propose an efficient and general solution that allows precise control over the camera when generating both photographic and artistic images. Unlike prior methods that rely on predefined shots, we rely solely on four simple extrinsic and intrinsic camera parameters, removing the need for pre-existing geometry, reference 3D objects, and multi-view data.
We also present a novel dataset with more than 57,000 images, along with their text prompts and ground-truth camera parameters. 
Our evaluation shows precise camera control in text-to-image generation, surpassing traditional prompt engineering approaches. Our data, model, and code are publicly
available at \url{https://graphics.unizar.es/projects/PreciseCam2024}.
\end{abstract}

\section{Introduction}
\label{sec:intro}

An image is a versatile medium whose content and camera language play essential roles. Images may be used as expressive art forms, where the same content but different camera parameters may convey dramatically different messages. For instance, a low camera angle makes a character dominating or a scene more epic, while a Dutch angle ($\approx{}\!45^{\circ}$ roll) induces a sense of uneasiness or tension.

Text-to-image (T2I) diffusion models can produce images with rich, varied content based on prompts.
However, these images often present a flat camera angle, i.e., the camera appears to have been placed parallel to the ground plane, with the horizon in the middle.
This poses a major limitation for creative expression, significantly reducing the potential for graphic designers, artists, and photographers to transmit emotion, sensations, or messages; to unlock and fully leverage the capabilities of generative models, the ability to precisely specify camera parameters is a must. 

Any particular camera view is defined by intrinsic and extrinsic parameters such as field of view (FoV), pitch, roll, etc.
However, these parameters are difficult to set through prompts since most test-to-image models have not been explicitly designed for camera view control and have not been trained on a diverse range of camera views. This leads to prompt engineering being the only tool to control camera view, requiring users to craft prompts precisely through trial and error to achieve the desired visual results. This not only demands a high level of expertise, but also offers a rather coarse and limited control.

\begin{figure*}[h]
  \centering
  \includegraphics[width=\textwidth]{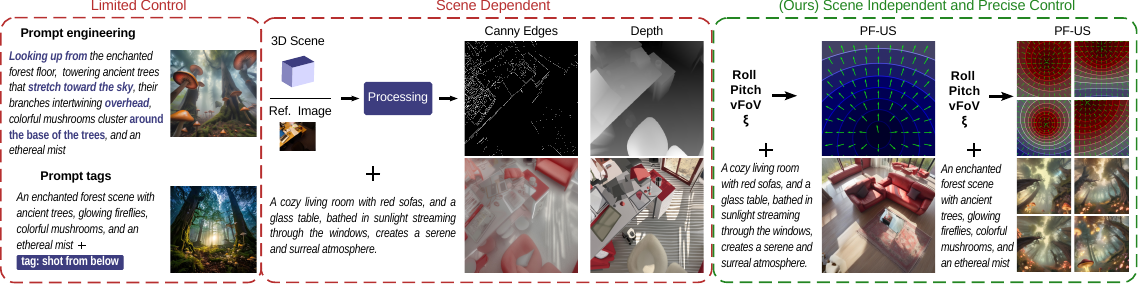}
  \caption{\textbf{Red boxes: }Current approaches rely on trial-and-error prompt engineering or generalistic tags, offering limited camera control in text-to-image generative AI. Others use 3D representations of the scenes, from which depth or edge maps are obtained, but this imposes strict constraints on the resulting image, limiting flexibility.  \textbf{Green box: }Our method relies on two extrinsic and two intrinsic camera parameters, user-provided; we then obtain the effects that such camera parameters impose on the appearance of each pixel in the final image, encoded in a 2D PF-US map (see text for details). This allows for precise, fine-tuned camera control, thus enhancing creativity. }
  \label{fig:comp_baselines}
\end{figure*}

Recently, diffusion models such as Firefly~\cite{firefly} have introduced some control over the camera view, allowing users to specify general instructions like \textit{wide angle}, \textit{shot from below/above}, or \textit{closeup}. 
However, these controls remain imprecise, keeping artistic options limited.
Precise camera view control has been explored by learning 3D information from multi-view images~\cite{kumari2024customizingtexttoimagediffusioncamera,cheng2024learning}, 
but such multi-view images are not always available, and learning 3D representations that cover all possible camera configurations is unattainable. 
As a result, these methods are limited in their ability to generate multiple objects, handle complex or cluttered scenes, and produce coherent backgrounds.
An alternative approach to 3D representations is to train ControlNet~\cite{zhang2023controlnet} on depth or edge maps, but unfortunately, these impose strict constraints beyond camera control, limiting flexibility. Fig.~\ref{fig:comp_baselines} (red boxes) summarizes these approaches.

In this work, we aim to provide a general solution to the problem of precise camera view control, which expands artistic freedom in generative AI. We do not focus solely on well-known photographic shots; instead, we offer simple, direct control of both intrinsic and extrinsic camera parameters, which allows us to achieve a wide spectrum of possible camera views while maintaining precise control (see Fig.~\ref{fig:comp_baselines} (green box)).
Instead of representing a 3D scene, which has the limitations we discussed above, we identify the essential effects that such camera parameters impose on the appearance of each pixel in the final image, thus overcoming the requirement to rely on multi-view images and using only simple, \textit{single-view} images.

To achieve our goal, we rely on two extrinsic parameters, roll and pitch rotations, and two intrinsic parameters, vertical FoV (vFoV) and geometrical distortion of the camera ($\xi$). Combined with a user-provided prompt, these parameters serve as inputs to our model to generate images with the intended camera view. We adopt the Unified Spherical (US) camera model~\cite{barreto2006unifying}, which allows us to translate these camera parameters into a Perspective Field (PF) representation~\cite{jin2023perspective}, which is easy to store and interpret. The PF-US encodes the camera parameters as per-pixel information, describing the pixels' up-vector (opposed to the gravity direction) and distance to the horizon, allowing us to encode per-pixel appearance from roll, pitch, vFoV, and $\xi$ parameters.
We then adopt ControlNet~\cite{zhang2023controlnet} to guide our image generation according to the resulting PF-US. Our model, \mname{}, archives precise camera view control, while the content remains fully determined by the text prompt.


In addition, we have created a novel dataset of 57,380 single-view RGB images, along with their ground-truth camera parameters and prompt descriptions. Our dataset spans a comprehensive range of camera parameters and a diversity of scenes, making it specifically suitable for the camera control problem we address. 

Last, while our approach is primarily designed for image generation, it opens up new possibilities for video generation, as absolute camera control conditioning is currently being overlooked. Our method can create a precise initial camera position, from which relative camera control can subsequently be applied. Alternatively, we showcase how our model tailored for images can be leveraged to apply limited absolute camera control on videos.
Our code, data, and model are publicly
available at \url{https://graphics.unizar.es/projects/PreciseCam2024}.

\section{Related Work}

\paragraph{Conditional image generation.}
Seminal work in conditional image generation leveraged local image statistics~\cite{imageanalogies,texsynthesis,imagequilting} and retrieving information from
large image datasets~\cite{sketch2photo,hays07,photosketch,semanticphotosynthesis}.
Later deep learning techniques allowed faster, more flexible, and more customized image synthesis within specific domains~\cite{spade,chen2017photographic,zhang2017stackgan,zhang2018stackgan++,wang2018high,wang2018video,isola2017image}.
Those models were limited in their ability to generate high-quality images and to support open-ended generation tasks.
With the advent of large-scale image and text datasets, auto-regressive~\cite{parti} and especially diffusion models~\cite{ldm,imagen,kandinsky} were proposed as an effective solution to create high-quality images from a virtually unbounded set of text descriptions. Despite these advancements, text remains a limited modality for controlling image generation.
Recent methods introduced modifications to the diffusion process, enhancing controllability by allowing users to incorporate additional guidance through color~\cite{sdedit}, hand-crafted terms~\cite{diffusionhandles,selfguidance}, and spatially localized prompts~\cite{multidiffusion,scenecomposer,ediffi,zeroshotlayout}.
While these methods offer improved control in the generative process, they lack support for more granular guidance,
such as structural cues like edges or depth maps.

\paragraph{Camera control on diffusion generation.}
While some works have addressed camera control for image generation, this area remains underexplored. Kumari et al.~\cite{kumari2024customizingtexttoimagediffusioncamera} proposed a method that introduces camera control without relying on prompt engineering. They fine-tune a text-to-image diffusion model, conditioning it on a 3D representation of an object learned in the model’s feature space, allowing for generating different camera viewpoints within new background scenes. However, their NeRF-based approach requires multi-view images of the objects and struggles with extreme camera angles or prompts involving multiple objects.
Cheng et al.~\cite{cheng2024learning} introduced Continuous 3D Words, a method for controlling attributes like object position, lighting, or some camera control (i.e., dolly zoom). Their approach disentangles these attributes from object identity by mapping them to the token embedding domain, enabling control during inference by integrating them in the prompt. However, it fails to follow complex prompts or specific styles and often resorts to object poses seen in the training set. 
We propose a general approach for generating images of any object, scene, or landscape while preserving the model's ability to handle complex prompts and produce various artistic styles and multi-object scenes. 

Closely related to our work are methods that train networks to guide the generative process using inputs like depth, edges, normals, and segmentation maps~\cite{zhang2023controlnet,t2iadapter}.
However, these methods still overlook a critical aspect of image creation: controlling the camera used to obtain that image.
It is common for practitioners to reason about image editing operations using camera abstractions -- what the image would look like if one changes the camera pitch, roll, field of view, and so on.
To address this gap, we propose a novel approach that incorporates fine-grained camera controls into diffusion adapters like ControlNet~\cite{zhang2023controlnet},
expanding the versatility and usability of image generation to scenarios where \emph{precise} camera control is paramount.
Last, recent works focus on relative camera view control for video generation~\cite{wang2024motionctrl,wang2024videocomposer,xu2024camco,he2024cameractrl,yin2023dragnuwa}. However, while these methods adjust camera movement relative to the first frame, they lack control over the initial camera position that sets the entire sequence. Our method can be directly applied to enable control over this starting camera position.
\section{Our Approach}
\label{sec:model}

Our proposed framework for precise camera view control in text-to-image generation takes as input a text prompt $p$ and a camera view specified by a set $\Omega$ of four parameters, 
and generates images $I$ according to both. 
The four camera parameters (roll, pitch, vFoV, and distortion $\xi$) are provided by the user through simple sliders; they are both intuitive and expressive enough to allow for precise camera control. These parameters,
together with the camera view representation used by our model, 
are presented in Sec.~\ref{subsec:PFrepresentation}.

Text-conditioned image generation is performed by means of a diffusion model $\mathcal{D}$,
while we achieve camera view conditioning via a ControlNet-based module, as described in Sec.~\ref{subsec:model}. The model is trained with a new dataset featuring ground-truth camera view specification (Sec.~\ref{subsec:dataset}).

\subsection{Precise Camera View Representation}
\label{subsec:PFrepresentation}

To enable a precise camera view specification, we adopt the Perspective Field (PF) representation~\cite{jin2023perspective}, which was originally designed for single-image camera calibration. For any arbitrary projection function $\mathcal{P}(\mathbf{X}) = \mathbf{x}$ mapping a 3D point $\mathbf{X}$ to an image pixel $\mathbf{x}$, PF assigns each pixel $\mathbf{x}$ an up-vector $\mathbf{u_{\mathbf{x}}}$ and a latitude angle $\varphi_{\mathbf{x}}$. Since each pixel $\mathbf{x}$ originates from a light ray $\mathbf{R}$ emitted from $\mathbf{X}$, the up-vector $\mathbf{u_{\mathbf{x}}}$ represents the projection of the up direction of $\mathbf{X}$ (which is opposite to the gravity vector $\mathbf{g}$ at $\mathbf{X}$), while the latitude $\varphi_{x}$ measures the angle between the light ray $\mathbf{R}$ and the horizontal world plane. In particular:
{\small
\begin{equation}
    \mathbf{u_{x}}=\lim _{c \rightarrow 0} \frac{\mathcal{P}(\mathbf{X}-c\mathbf{g})-\mathcal{P}(\mathbf{X})}{\|(\mathcal{P}(\mathbf{X})-c\mathbf{g})-\mathcal{P}(\mathbf{X})\|_2}, \, \textrm{and}
    \label{eq:upVector}
\end{equation}
}
{\small
\begin{equation}
    \varphi_{\mathbf{x}}=\arcsin \left(\frac{\mathbf{R} \cdot \mathbf{g}}{\|\mathbf{R}\|_2}\right).
\end{equation}
}

\begin{figure*}[h]
  \centering
  \includegraphics[width=\textwidth]{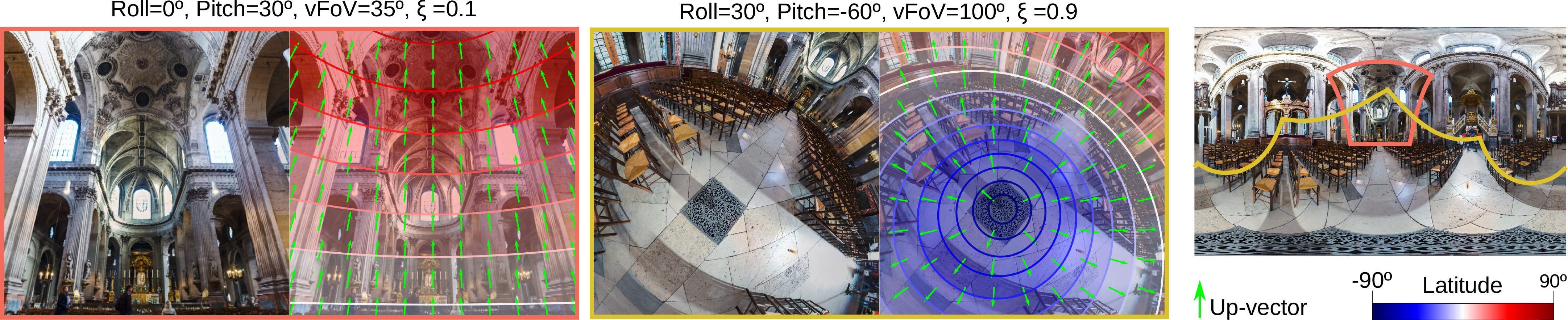}
  \caption{
  \textbf{Left and center: PF-US camera view representation.}
  PF-US camera parameters (roll, pitch, vertical FoV, and distortion $\xi$) and associated maps corresponding to two example images. PF-US maps encode, for each pixel, a latitude value $\mathbf{\varphi_x}$ (blue-red color map with contour lines) and a unit up-vector $\mathbf{u_x}$ (green arrows). 
  \textbf{Right: Dataset generation.} 
  We generate training data in the form of images with ground-truth PF-US maps by leveraging 360$^\circ$ images.  
  We sample our camera parameters and 
obtain the corresponding cropped region for each sampled quartet (as examples, the highlighted regions in red and yellow yield the two examples shown on the left).  
  }
  \label{fig:pf_expl}
\end{figure*}

We aim to enhance creativity and expressivity beyond the commonly used pinhole model in generative AI, including precisely-controlled wide-angle or fisheye cameras. To achieve this, we take advantage of the flexibility of PF to support different camera models through their respective $\mathcal{P}(X)$ functions, and employ the Unified Spherical (US) camera model~\cite{barreto2006unifying,bogdan2018deepcalib}, which allows camera models beyond the pinhole camera 
to be encoded within the PF. 
The projection function $\mathcal{P}(X)$ of the US model is defined as:
{\small
\thinmuskip=\muexpr\thinmuskip*5/8\relax
\medmuskip=\muexpr\medmuskip*5/8\relax  
\begin{multline}
    \mathcal{P}(\mathbf{X}) = \mathbf{x} = (u,v),\\
    u = \frac{x f}{\xi \sqrt{x^{2}\hspace{-.2em}+\hspace{-.2em}y^{2}\hspace{-.2em}+\hspace{-.2em}z^{2}}\hspace{-.2em}+z}+u_0,
    v =\frac{y f}{\xi \sqrt{x^{2}\hspace{-.2em}+\hspace{-.2em}y^{2}\hspace{-.2em}+\hspace{-.2em}z^{2}}\hspace{-.2em}+z}+v_{0}.\\
\raisetag{20pt}\label{eq:US}
\end{multline}
}
where $(x,y,z)$ are the 3D point world coordinates, $(u_{0},v_{0})$ are the principal point coordinates in the image, $f$ is the focal length, and $\xi$ is a distortion parameter ranging between 0 and 1, with $\xi=0$ being a pinhole camera. 

We choose to represent our camera view with two intrinsic parameters: vertical FoV (which is in turn defined by the focal length $f$) and distortion $\xi$, as well as two extrinsic parameters: roll and pitch rotations with respect to the horizon. These four parameters conform the set $\Omega = (\textrm{roll}, \textrm{pitch}, \textrm{vFoV}, \xi)$, which is expressive enough to represent a wide range of camera configurations, while being intuitive to users.
The four parameters are then transformed, by means of the PF-US representation (Eqs.~\ref{eq:upVector}-\ref{eq:US}), into per-pixel up-vectors and latitude values, yielding the corresponding PF-US maps (see Fig.~\ref{fig:pf_expl}, left and center, for two different illustrative examples). 

The resulting PF-US maps provide \emph{local} information about how camera parameters affect the appearance of each pixel, allowing the model to learn this relationship without the need for heavy 3D representations.
Finally, note that yaw rotation is excluded; given a 2D image, we can define up or down directions based on the horizon, but there is no such reference to define left or right directions. By leaving yaw information out, we allow the model to focus on the relevant camera parameters that affect image formation. 

\begin{figure}[t]
  \centering
  \includegraphics[width=\columnwidth]{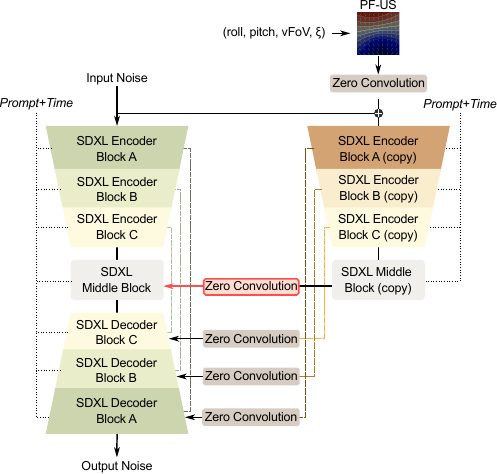}
  \caption{
  Training of our proposed approach for precise camera control. Text-to-image generation with a UNet-based diffusion model (SDXL in our implementation) is conditioned with a PF-US map representing the desired camera parameters; conditioning is learned by means of a ControlNet-based module. 
  During inference, only the middle block output of this module is injected into SDXL (shown in red, see text for more details).}
  \label{fig:model}
\end{figure}

\subsection{Learning Camera View Control}
\label{subsec:model}

We aim to introduce precise camera view control in text-to-image diffusion models, while preserving their generalization capabilities. This requires a method that minimally disrupts their generation pipeline, already trained on large, quality datasets, adding only the essential information needed to achieve the desired camera view. 
To do this, we adopt the ControlNet approach to condition image generation~\cite{zhang2023controlnet}: it provides a framework for guiding diffusion models to adhere to a specific condition, and has delivered impressive results conditioning generation with poses, depth, edges, or normal maps, among others. 

During the generation process, ControlNet ensures that the generated images align with both the prompt and the conditioning input, in our case the PF-US map. 
We use Stable Diffusion XL (SDXL)~\cite{podell2023sdxl} as our base text-to-image model, although our approach is compatible with any other UNet-based diffusion model.
Our ControlNet setup involves duplicating the encoder and middle block layers of the original SDXL model. While SDXL remains frozen during training to preserve its generalization abilities, ControlNet is initialized with the same weights and trained (refer to the \supplem{supplementary material} for training details). 

Fig.~\ref{fig:model} illustrates this. To incorporate ControlNet's output into SDXL, its residuals are passed through a zero-convolution layer (1×1 convolution layer with all-zero initial weights) before being added to the residuals of the original SDXL model. 
During training, outputs from ControlNet are added both to the U-Net bottleneck and to the decoder skip connections, as in the original implementation. At inference time, however, we find that injecting the ControlNet output \textit{only} in the bottleneck (marked in red in Fig.~\ref{fig:model}) improves generation consistency without being detrimental to condition adherence~\cite{bhat2024loosecontrol}.
Our ablation study in Sec.~\ref{sec:results} illustrates the effect of the different ControlNet outputs on the generation process.

\subsection{Dataset Generation}
\label{subsec:dataset}

To train our proposed framework, we need a dataset of RGB images and their corresponding text prompts and PF-US camera parameters, i.e., triplets ($I_i$, $p_i$, $\Omega_i$). We need this dataset to be diverse in content, as well as covering a wide range of camera parameters.

Jin et al.~\cite{jin2023perspective} provide data featuring RGB images and ground-truth PF representations. However, they primarily depict urban outdoor scenes and, more importantly, do not cover the full range of camera parameters; for example, they include only minimal distortions (low $\xi$ values) and avoid large vertical FoVs. 
A possible alternative would be to use their PF estimation model and apply it to an existing dataset. 
However, the resulting estimations lack sufficient accuracy for our training purposes, hindering our model's ability to learn view control across the full spectrum of camera parameters (see \supplem{supplementary material}). 

We thus generate our own dataset with ground truth PF-US representations by leveraging 360$^\circ$ images. We sample our set of camera parameters (see \supplem{supplementary material} for more details) and obtain, for each sampled quartet, the corresponding areas cropped from the 360$^\circ$ images, and their PF-US maps (as described in Sec.~\ref{subsec:PFrepresentation}). 
Fig.~\ref{fig:pf_expl} (right) shows two of the resulting regions, with boundaries highlighted in orange and yellow, respectively. 
Sampling ranges for each camera parameter are: roll $\in (-90^\circ,90^\circ)$, pitch $\in (-90^\circ,90^\circ)$, vFoV $\in [15^\circ,140^\circ]$ and $\xi \in (0, 1)$. 

To maximize diversity in the images, we use six different 360$^\circ$ image datasets: 360-SOD~\cite{li2019_360SOD}, CVRG-Pano~\cite{orhan2022CVRGPano}, F-360iSOD~\cite{zhang2020F-360iSOD}, Poly Haven HDRIs~\cite{polyhaven}, Sitzmann et al.~\cite{sitzmann2018saliency}, and 360cities~\cite{360cities}. These feature a mix of outdoor and indoor scenes, containing both natural and urban settings with a diverse array of activities and environments. 

From the resulting ($I$, $\Omega$) pairs, we use BLIP-2~\cite{li2023blip} to generate a descriptive text prompt $p_i$ for each image $I_i$. 
While BLIP-2 may sometimes provide inaccurate descriptions, this is not an issue for our training: we do not train the text-to-image model itself, but the conditioning module instead, which needs to learn the camera view independently of the prompt. These inaccurate prompts are akin to the ControlNet technique of introducing empty strings as the prompt during training.
Our final dataset is comprised of 57,380 RGB images with associated text prompts and ground-truth PF-US parameters. 

\section{Results}
\label{sec:results}
\begin{figure*}[t]
  \centering
  \includegraphics[width=1\textwidth]{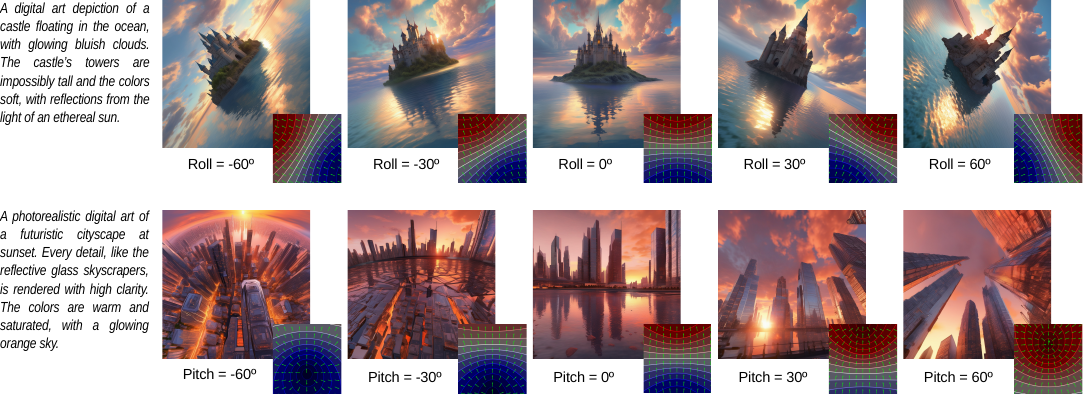}
  \caption{
  \textbf{Extrinsic Parameter Control.} Images generated by varying the extrinsic camera parameters while keeping the rest of the conditioning fixed, showing consistent adherence to the camera specification. Top row: Variation in pitch, effectively shifting the view from looking downward to upward; fixed parameters are $(\textrm{roll}, \textrm{vFoV}, \xi)$ = (0º, 80º, 0.1). Right: Variation in roll, tilting the view from left to right; $(\textrm{pitch}, \textrm{vFoV}, \xi)$ = (0º, 80º, 0.1). Insets show the corresponding PF-US maps, and text prompts are shown in italicized text.
  }
  \label{fig:exparam_ctl}
\end{figure*}
\begin{figure*}[t]
  \centering
  \includegraphics[width=1\textwidth]{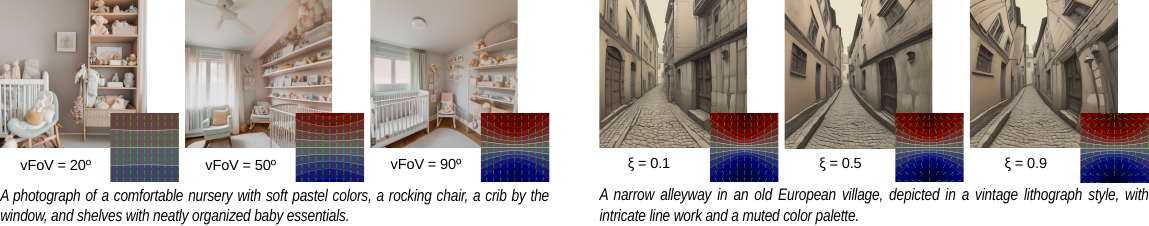}
  \caption{
    \textbf{Intrinsic Parameter Control.} Images generated by varying the intrinsic camera parameters while keeping the rest of the conditioning fixed, showing consistent adherence to the camera specification. Left: Variation in the vertical field of view (vFoV); fixed parameters are $(\textrm{roll}, \textrm{pitch}, \xi)$ = (0º, 0º, 0.1). Right: Variation in distortion ($\xi$); $(\textrm{roll}, \textrm{pitch}, \textrm{vFoV})$ = (0º, 0º, 120º). Insets show the corresponding PF-US maps, and text prompts are shown in italicized text.
    }
  \label{fig:intparam_ctl}
\end{figure*}

This section provides an evaluation of \mname{}'s performance. We illustrate the precise control of camera parameters, followed by a comparison with baseline methods, and show that our method can handle both artistic and realistic styles. We also study the robustness of our model, analyzing camera conditioning adherence and conducting an ablation study on residuals to refine control and quality. Finally, we showcase various applications of our method, including background generation for object rendering, video generation, and extended control with multiple ControlNets, to illustrate its versatility. The \supplem{supplementary material} provides additional results and details.

\paragraph{Camera control.} 
\mname{} enables precise control over extrinsic (roll and pitch rotations) and intrinsic (vFoV and distortion $\xi$) camera parameters. 
Fig.~\ref{fig:exparam_ctl} illustrates control over extrinsic parameters. In it, roll (top row) and pitch (bottom row) are varied while the remaining parameters are held constant. The images follow the conditioning, and a high degree of consistency is maintained through camera variations.
Similarly, in Fig.~\ref{fig:intparam_ctl} we show intrinsic parameter control by varying the vertical field of view vFoV (left) or the distortion $\xi$ (right), while keeping all the other parameters fixed. In the PF-US maps (see Sec.~\ref{subsec:PFrepresentation}), larger vFoV leads to increased latitude values, whereas $\xi$ mainly affects the up vectors. The generated images follow the PF-US maps, effectively translating camera settings into visual outcomes.
The \supplem{supplementary material} includes more examples of systematic variation of camera parameters for a variety of prompts.

\begin{figure*}[t]
  \centering
  \includegraphics[width=0.95\textwidth]{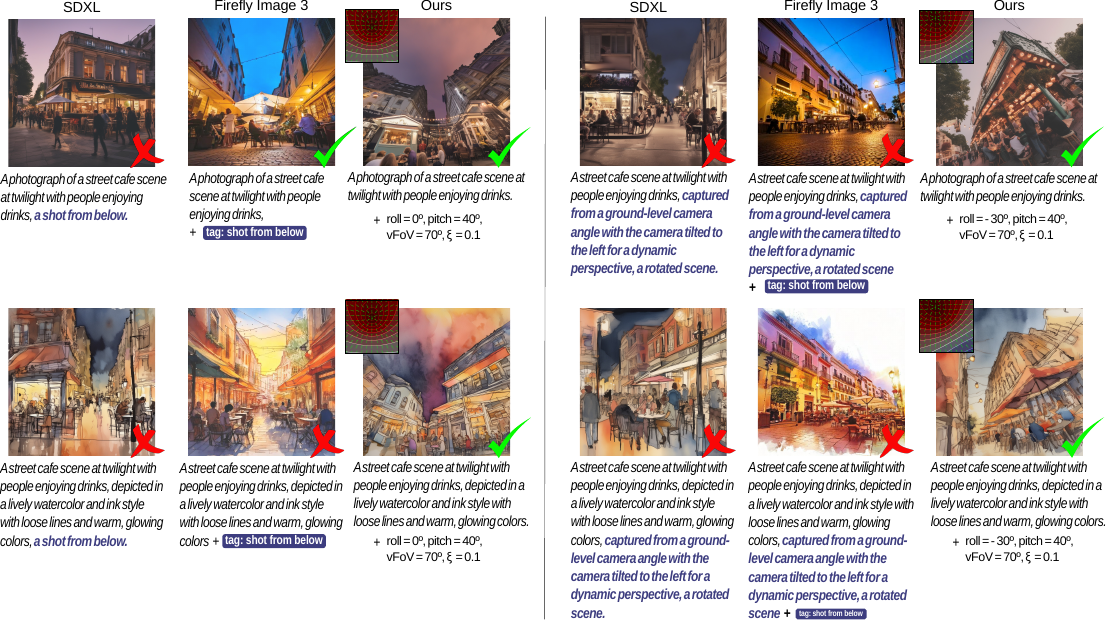}
  \caption{
  Comparison of our model with SDXL and Adobe Firefly (Firefly tags are in dark blue background; prompt engineering, when present, in bold and dark blue text). Four examples are shown: the top half depicts realistic styles, while the bottom half depicts artistic styles; the left half examples intend a \emph{shot from below}, and the right half a \emph{tilted view from below}. SDXL struggles in all cases. Firefly successfully produces the \emph{shot from below} in the realistic style due to its matching tag, but fails in the artistic version of it; for more complex perspectives, like the \emph{tilted view from below}, Firefly lacks a tag for this view, and fails to produce a valid result despite the efforts with the prompt. Our approach achieves reliable camera control across both styles and perspectives.
  }
  \label{fig:comparative_T2I}
\end{figure*}

\begin{figure}[h]
  \centering
  \includegraphics[width=\columnwidth]{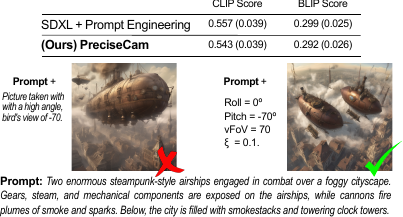}
  \caption{
  Our model achieves prompt adherence on par with SDXL, as reflected in the table by comparable CLIP and BLIP scores (see text for details), while enabling camera control that SDXL cannot provide. Images show an example result from SDXL with prompt engineering (left), and from our model (right). 
  }
  \label{fig:scores}
\end{figure}

\paragraph{Comparisons.} 
We compare our approach to two alternative methods: prompt-engineered SDXL and Adobe Firefly, a state-of-the-art diffusion model with preset style tags. Fig.~\ref{fig:comparative_T2I} illustrates this comparison. Despite prompt engineering efforts, SDXL’s ability to control camera perspectives remains limited, often struggling to consistently interpret complex camera-related prompts. Adobe Firefly includes predefined style tags for shot specification, such as \textit{close-up}, \textit{shot from above}, \textit{shot from below}, and \textit{wide angle}, supporting basic camera control. Nevertheless, while Firefly’s tags do introduce some perspective variations, the control enabled by these tags is limited, and insufficient even when combined with prompt specification. In contrast, our model is able to control camera views precisely in both realistic (top row) and artistic (bottom row) styles.
We also evaluate whether our model maintains, despite the inclusion of camera control, the text prompt adherence exhibited by SDXL. To do this, we compare our results to those of the baseline SDXL by computing, for each method, CLIP and BLIP scores~\cite{hessel2021clipscore}, which are commonly used to evaluate prompt relevance in generated images. The comparison is done over 2,940 images generated with each method (ours and baseline SDXL). For the baseline SDXL, we use prompt engineering, specifying the desired view to encourage distinct camera perspectives (see the \supplem{supplementary material} for more details). Our approach achieves comparable scores to SDXL (Fig.~\ref{fig:scores} shows mean and standard deviation), suggesting that our model is able to maintain prompt alignment while providing precise camera view control. 

\subsection{Model analysis}

\paragraph{Camera conditioning adherence.} 
To assess the stability of our model’s camera conditioning, we generate images using the same PF-US maps and prompts but vary the input noise. Results show consistent camera conditioning
(please refer to the \supplem{supplementary material} for examples).

\begin{figure}[t]
  \centering
  \includegraphics[width=0.9\columnwidth]{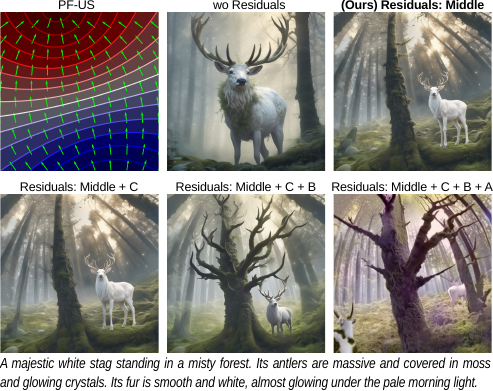}
  \caption{
  Image generation conditioned by the text prompt (bottom) and fixed camera parameters, for different injections of residual outputs of the ControlNet-based module into the main SDXL model. 
  Middle block residuals injected at the bottleneck (ours) enable precise camera control with minimal impact on image quality, while deeper residuals introduce distortions, reducing overall image quality. 
  }
  \label{fig:ablation-residuals}
\end{figure}

\vspace{-10px}
\paragraph{Ablation: Influence of the residuals' contributions.}
In our framework, the ControlNet-based module generates residual outputs that can be injected into the main model in various layers. We observe that residuals from the middle block (bottleneck) effectively achieve adherence to the camera conditioning, while preserving the image quality achieved by the base model (SDXL). The injection of residuals at additional levels can introduce distortions and hinder semantic integrity preservation, as shown in Fig.~\ref{fig:ablation-residuals}. This aligns with the findings of previous works that use ControlNet for conditioning diffusion models~\cite{bhat2024loosecontrol}. 
Therefore, in our model we use only the mid-level residuals during inference; these are key for adhering to camera conditions without disrupting the generative capabilities of the base model (SDXL), providing a good trade-off between effective camera control and high-quality image generation.

\subsection{Applications}

\begin{figure}[t]
  \centering
  \includegraphics[width=\columnwidth]{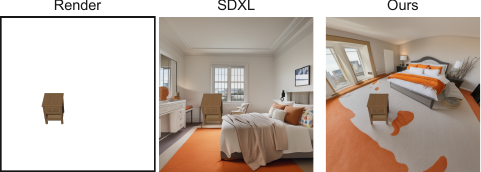}
  \caption{Background generation without and with camera control. Our method can generate a background that aligns with the object’s perspective (shown on the left), ensuring seamless integration. The baseline approach (SDXL, center) lacks perspective consistency, while the background generated with our method aligns well with the object (right).
  }
  \label{fig:obj_back}
\end{figure}

\paragraph{Background generation for object rendering.}
\mname{} can be used to generate backgrounds that match the perspective of a given object. By aligning the background’s perspective with the object’s viewpoint, our model can be used to create visually coherent scenes where the object appears naturally embedded, as shown in Fig.~\ref{fig:obj_back}.

\begin{figure}[t]
  \centering
  \includegraphics[width=\columnwidth]{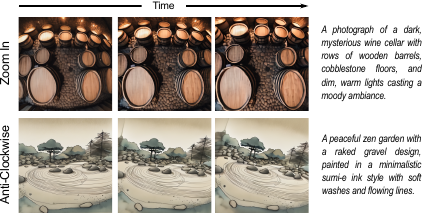}
  \caption{Video generation results showing two examples of frame-by-frame camera control. Camera parameters are specified to achieve a zoom-in effect (top) and an anti-clockwise rotation (bottom). Frames progress from left to right over time.}
  \label{fig:video_results}
\end{figure}

\vspace{-10px}
\paragraph{Video generation.} Our model can be used to condition the camera view on each frame of a video to specify the desired perspectives throughout the sequence, as illustrated in Fig.~\ref{fig:video_results}. Each frame is guided by our four camera parameters through a custom-trained adapter~\cite{lin2024ctrladapter} for Stable Video Diffusion~\cite{blattmann2023SVD}, ensuring consistent perspective control across frames. Alternatively, our method can be used to condition the initial frame of a video sequence, after which existing methods for relative camera view control can guide the trajectory for the remaining frames~\cite{wang2024motionctrl,wang2024videocomposer,xu2024camco,he2024cameractrl,yin2023dragnuwa}. While these existing methods establish movement relative to the first frame, they lack control over the initial camera position that sets the starting perspective for the entire video. Our approach addresses this gap by enabling precise control over the initial camera position to anchor the video sequence.

\paragraph{Extended control with multiple ControlNets.} Additionally, our approach is compatible with other ControlNets, allowing additional conditioning on the final output to achieve complex effects. For instance, combining our model with depth, Canny edges, and pose control can allow for simultaneous control over camera view, subject positioning, and scene structure~\cite{zhang2023controlnet}. For further details and examples, please refer to the \supplem{supplementary material}.

\section{Conclusion}
\label{sec:conclusions}
We have presented a framework to enable precise camera control for text-to-image diffusion-based generation. 
We have also provided a novel dataset tailored to this problem. Our experiments demonstrate precise and consistent control of the camera, far exceeding what is possible with current approaches, including prompt engineering. 
Our model works even with paintings or artistic images (e.g., Fig.~\ref{fig:comparative_T2I}), despite the absence of this type of images in our dataset. 
Moreover, we have also shown proof-of-concept examples of applications beyond enhancing creative expression, including background generation for object compositing, and video generation. 

\begin{figure}[t!]
  \centering
  \includegraphics[width=\columnwidth]{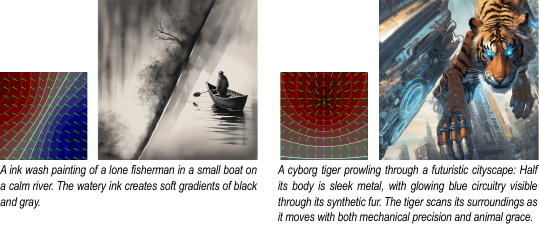}
  \caption{\textbf{Limitations.} Left: The model may on some occasions incorrectly position objects or people in a vertical, standing orientation (like the man on the boat), as it is more frequent.
  Right: In the presence of conflicts between the prompt and the desired camera view, the model may produce semantically incoherent outputs.
  }
  \label{fig:limitations}
\end{figure}

Although our dataset covers a wide range of camera parameters and content, 
\mname{} may struggle with extreme roll rotations, since the SDXL backbone tends to align objects vertically (Fig.~\ref{fig:limitations}, left). Also, while our model retains SDXL’s ability to generate images matching the prompt, conflicts between the prompt description and the specified camera view (Fig.~\ref{fig:limitations}, right) may lead to incoherent outputs. We hope our model and dataset become helpful in developing and perfectioning novel tools for artistic control of AI-generated images.

\section*{Acknowledgements}

We extend our gratitude to the members of the Graphics and Imaging Lab for their insightful discussions and assistance in preparing the final figures, with special thanks to Dario Lanza and Nestor Monzon. This research was funded by grant PID2022-141539NB-I00, funded by MICIU/AEI/10.13039/501100011033 and by ERDF, EU. Additionally, Edurne Bernal-Berdun was supported by the Gobierno de Aragon predoctoral grant program (2021–2025).

\clearpage
\setcounter{page}{1}
\setcounter{section}{0}
\renewcommand\thesection{\Alph{section}}
\maketitlesupplementary

\noindent The supplementary material for \textit{\mname{}: Precise Camera Control for Text-to-Image Generation} includes this PDF document, an HTML browser featuring additional image results, and a demo video showing the usability of our model.

\section{Training Details}
To train our framework we use the ControlNet~\cite{zhang2023controlnet} loss function. PF-US maps are encoded as RGB images, where the up-vector coordinates are scaled from [-1, 1] to [0, 255] and assigned to the R and G channels, while latitude values are mapped from [-90, 90] to [0, 255] and represented in the B channel. We initialize ControlNet using the SDXL model weights from Stability AI\footnote{\url{https://huggingface.co/stabilityai/stable-diffusion-xl-base-1.0}}, and train it on our entire dataset using the Adam optimizer~\cite{Kingma2014Adam} with hyperparameters $\beta_{1}=0.9$, $\beta_{2}=0.999$, weight decay $w=10^{-2}$, and a learning rate of 10$^{-6}$. We employ a total batch size of 32, an input resolution of 1024x1024 pixels, 16 floating-point precision, and 70,000 steps. As per usual practice, 50$\%$ of the text prompts are replaced by empty strings during training. The training was executed within the Accelerate framework~\cite{accelerate} for four days on eight NVIDIA RTX A100 GPUs.

\section{Dataset Details}

\begin{figure}[t]
  \centering
  \includegraphics[width=\columnwidth]{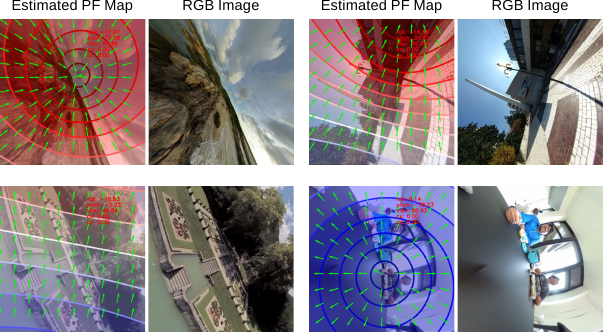}
  \caption{Incorrect PF map estimations using the model from Jin et al.~\cite{jin2023perspective} for different RGB images. These errors make the estimator unsuitable for our dataset creation, as the introduced noise is substantial enough to compromise our model's training.}
  \label{fig:pf_estimations}
\end{figure}

To train our model, we require triplets of RGB images, corresponding text prompts, and PF-US camera parameters ($I_i$, $p_i$, $\Omega_i$). It is essential for our dataset to be diverse in both content and camera parameter values. We explore several approaches:

\begin{itemize}
    \item \textit{Existing Datasets:} Jin et al.~\cite{jin2023perspective} present a dataset containing RGB images paired with ground-truth PF maps. However, they primarily depict urban outdoor scenes and lack comprehensive coverage of camera parameters. For instance, images depicting large vertical FoVs or extreme distortions are absent.
    
    \item \textit{PF estimators:} Previous works offer deep-learning models to estimate the PF map of a given image~\cite{jin2023perspective,veicht2024geocalib}, primarily intended for camera calibration. Thus, an alternative approach might be to apply this model to an existing image dataset, thus obtaining its associated PF maps. However, the estimated PF maps lack the precision required for our training needs, limiting our model's ability to learn effective camera view control across the full range of camera parameters (see Fig.~\ref{fig:pf_estimations}). Moreover, PF estimation models do not always consider the distortion parameter $\xi$.
    
    \item \textit{Cropping 360$^\circ$  images: } By using 360$^\circ$ images, we can extract patches corresponding to specific camera parameters, providing ground-truth PF-US maps that are crucial for our application. This approach allows us to sample the entire range of camera parameters while leveraging the Unified Spherical camera model, including its $\xi$ distortion parameter.
\end{itemize}

To generate our dataset with ground-truth PF-US maps, we adopt this last approach using 360$^\circ$ images. We sample our set of camera parameters and obtain, for each sampled quartet $\Omega$=(roll, pitch, vFoV, $\xi$) the corresponding patches cropped from the 360$^\circ$ images, and their PF-US maps. To maximize content diversity, we use six different 360$^\circ$ image datasets: 360-SOD~\cite{li2019_360SOD}, CVRG-Pano~\cite{orhan2022CVRGPano}, F-360iSOD~\cite{zhang2020F-360iSOD}, Poly Haven HDRIs~\cite{polyhaven}, Sitzmann et al.~\cite{sitzmann2018saliency}, and 360cities~\cite{360cities}. These feature outdoor and indoor scenes, containing both natural and urban settings with diverse activities and environments. 

From each 360$^\circ$ image, we sample 24 patches. 
To maximize the content diversity that each 360$^\circ$ image has to offer and avoid repeatedly sampling the same areas, the image is divided into six regions, with four patches sampled from each region using different camera parameters $\Omega$. 
For each region, we randomly sample yaw (necessary only to establish the 360$^\circ$ image horizontal coordinate) and pitch.
For each pair of yaw and pitch, we randomly sample two vFoV values (one small $\in (0, 0.5)$ and one large $\in [0.5, 1)$), two $\xi$ values (low $\in [15, 60)$ and high $\in [60, 140)$), 
yielding four possible combinations. We sample a roll rotation for each combination to generate four distinct image crops of the same region.
This approach ensures that the same image content is depicted across different image crops, showcasing both minimum and maximum vFoVs and varying distortion levels at different rotations. This allows the model to learn how these parameters influence the final image content (e.g., how the appearance of a chair at a high vFoV varies when $\xi$ is increased or decreased). 

This results in a dataset of 57,380 RGB images with a ground-truth PF-US condition. Sampling ranges for each camera parameter are: roll $\in (-90^\circ,90^\circ)$, pitch $\in (-90^\circ,90^\circ)$, vFoV $\in [15^\circ,140^\circ]$ and $\xi \in (0, 1)$. We use BLIP-2~\cite{li2023blip} to generate a descriptive text prompt $p_i$ for each image $I_i$. 

\section{Additional Results}
We present additional results of \mname{} for various prompts in the form of an HTML browser. We show how our model can accurately generate images with the specified camera view. Within each tab, we display in each row the generated images for the same prompt when a single camera parameter is varied while keeping the others fixed. The first row of each tab shows the PF-US for the corresponding camera settings. Note that the quality of the images has been reduced to meet the upload size in the paper submission platform.

\section{Prompt Engineering for Baseline SDXL}

In Sec.~\ref{sec:results} and Fig.~\ref{fig:comp_baselines}, we show how our model maintains the text prompt adherence exhibited by the baseline SDXL despite the inclusion of camera control, achieving comparable CLIP and BLIP scores~\cite{hessel2021clipscore}. This comparison is based on 2,940 images generated by both our method and the baseline SDXL. To enable the baseline SDXL to produce the correct camera views, we employed prompt engineering, explicitly specifying the desired view to encourage distinct camera perspectives.

This section outlines the prompt engineering techniques applied to SDXL. After extensive testing, we identified the following prompt engineering scheme as the most effective, occasionally producing camera views resembling the specified parameters. The focus was on roll, pitch, and vFoV, as distortion effects could not be replicated. To each prompt, we appended the following descriptions:

\begin{itemize}
    \item Roll below 0º: Dutch angle shot, frame tilted $<$\textit{roll value}$>$ degrees to the left.
    
    \item Roll above 0º: Dutch angle shot, frame tilted $<$\textit{roll value}$>$ degrees to the right.
    
    \item Pitch below 0º: Picture taken with a high angle, bird's view of $<$\textit{pitch value}$>$.
    
    \item Pitch above 0º: Picture taken with a low angle, worm's view of $<$\textit{pitch value}$>$.

    \item VFoV below than 30º: Picture taken with a vertical field of view of $<$\textit{vFoV value}$>$ degrees, a extreme close-up shot.

    \item VFoV between 30º and 55º: Picture taken with a vertical field of view of $<$\textit{vFoV value}$>$ degrees, a close-up shot.

    \item VFoV between 55º and 75º: Picture taken with a vertical field of view of $<$\textit{vFoV value}$>$ degrees, a medium shot.

    \item VFoV between 75º and 90º: Picture taken with a vertical field of view of $<$\textit{vFoV value}$>$ degrees, a long shot.

    \item VFoV above 90º: Picture taken with a vertical field of view of $<$\textit{vFoV value}$>$ degrees, a extreme long shot.
    
\end{itemize}

\begin{figure*}[h]
  \centering
  \includegraphics[width=0.8\textwidth]{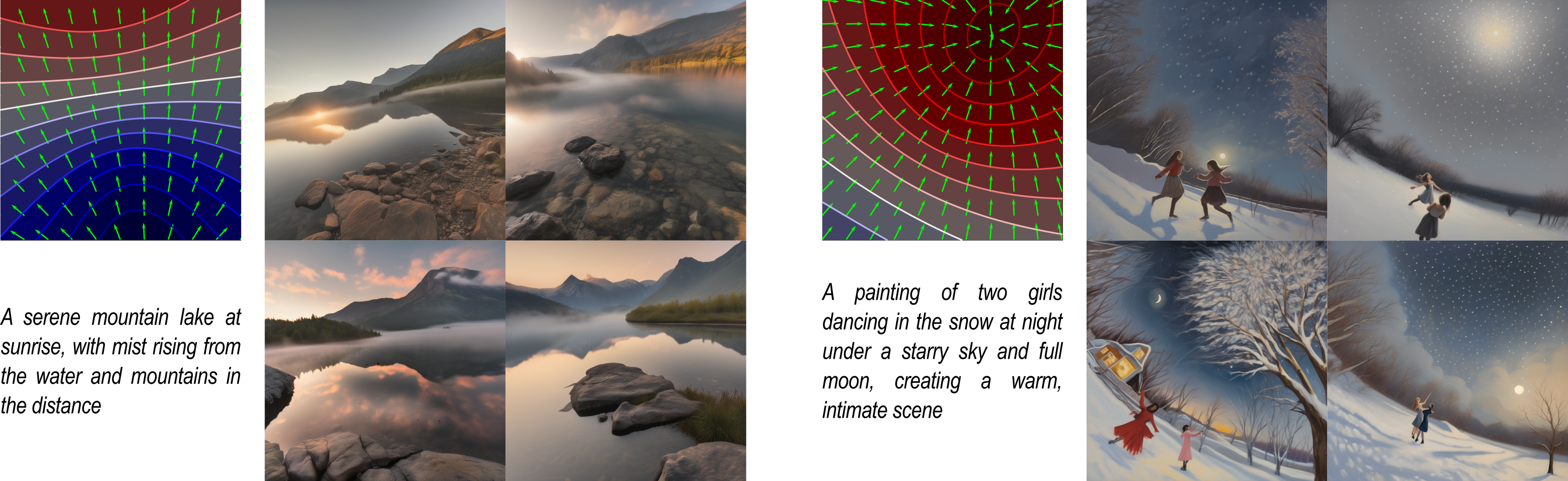}
  \caption{Generated images for different input noises but using the same prompt and camera parameters. \mname{} produce different images while adhering to the specified camera parameters represented as the PF-US map.
  }
  \label{fig:seed_consistency}
\end{figure*}

\begin{figure*}[h]
  \centering
  \includegraphics[width=0.8\textwidth]{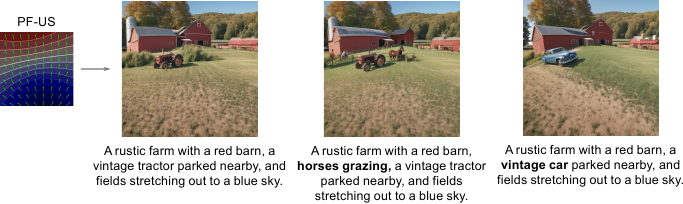}
  \caption{Generated images for small variations in prompt using the same noise and camera parameters. \mname{} produce different images based on the prompt description while maintaining the specified camera view represented as the PF-US map.}
  \label{fig:prompt_consistency}
\end{figure*}

\section{Consistent Camera for Input Variations}

\mname{} consistently generates the specified camera view regardless of variations in input noise or prompts. Fig.~\ref{fig:seed_consistency} shows our model's ability to produce diverse image alternatives with the correct camera view when the input noise varies while keeping the prompt and camera parameters fixed. Additionally, Fig.~\ref{fig:prompt_consistency} illustrates that changes in the prompt do not affect the final camera view of the generated images for a given set of camera parameters and input noise. Instead, the model adjusts the content to align with the modified prompt.

\section{Compatibility with Multiple ControlNets}

\begin{figure}[h]
  \centering
  \includegraphics[width=\columnwidth]{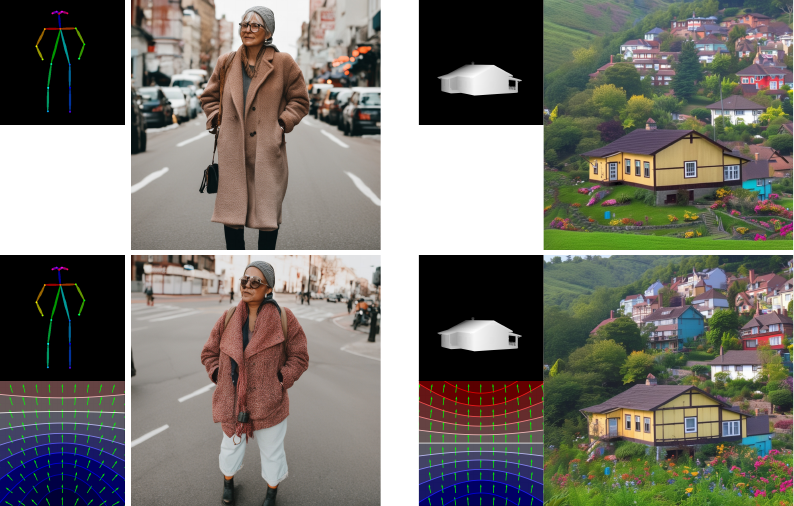}
  \caption{\mname{} is compatible with previous ControlNets, including pose control (left) and depth control (right). We showcase control over the person's pose while simultaneously controlling the camera view, and our ability to generate images based on depth inputs while maintaining a background with consistent perspective. In the depth example, observe the change in perspectives of the red house in the background when we include camera control (right-bottom).
  }
  \label{fig:multiCNet}
\end{figure}

\mname{} is compatible with other ControlNet models~\cite{zhang2023controlnet}, such as pose, depth, or edge maps. As shown in Figure~\ref{fig:multiCNet}, our model integrates seamlessly with various ControlNets. While pose control\footnote{\url{https://huggingface.co/thibaud/controlnet-openpose-sdxl-1.0}} adjusts the subject's pose, it does not control the background. By using \mname{}, we apply camera view control to the final image while achieving the desired person's pose. Additionally, in challenging cases where depth maps only represent objects without defining the background's depth, our model can boost the generation of a coherent background with an accurate perspective\footnote{\url{https://huggingface.co/diffusers/controlnet-depth-sdxl-1.0}}. Notice in Fig.~\ref{fig:multiCNet} how the background perspective generated with \mname{} aligns more closely with the house's perspective.

\section{Video Demo}
We provide a supplementary video highlighting the usability of our model. The video shows how users can intuitively adjust camera parameters with sliders to preview the desired camera view and generate an image based on the provided prompt.

\clearpage
 
{
    \small
    \bibliographystyle{ieeenat_fullname}
    \bibliography{main}
}

\end{document}